# A Comparative Study of Filtering Approaches Applied to Color Archival Document Images

Walid Elhedda*, Maroua Mehri and Mohamed Ali Mahjoub

LATIS - Laboratory of Advanced Technology and Intelligent Systems, Sousse University,
National Engineering School of Sousse, 4023, Sousse, Tunisia

**Abstract:** *Current systems used by the Tunisian national archives for the automatic transcription of archival documents are hindered by many issues related to the performance of the optical character recognition (OCR) tools. Indeed, using a classical OCR system to transcribe and index ancient Arabic documents is not a straightforward task due to the idiosyncrasies of this category of documents, such as noise and degradation. Thus, applying an enhancement method or a denoising technique remains an essential prerequisite step to ease the archival document image analysis task. The state-of-the-art methods addressing the use of degraded document image enhancement and denoising are mainly based on applying filters. The most common filtering techniques applied to color images in the literature may be categorized into four approaches: scalar, marginal, vector and hybrid. To provide a set of comprehensive guidelines on the strengths and weaknesses of these filtering approaches, a thorough comparative study is proposed in this article. Numerical experiments are carried out in this study on color archival document images to show and quantify the performance of each assessed filtering approach.*

## 1. Introduction

In order to strengthen initiatives in the exploitation and enrichment of the national documentary heritage, the Tunisian national archives[1] are extremely interested in providing reliable tools for assisting researchers and historians in manual transcription of the archival documents. The Tunisian national archives digitize over 200 pages per day. The documentary heritage digitized by the Tunisian national archives encompasses more than six centuries (1500-2000) of Tunisian history. It consists primarily of printed and manuscript image documents written in Arabic and Latin and digitized at 300 dpi in full color mode. Analyzing historical documents by using a classical OCR system requires burdensome and complex processing due to the idiosyncrasies and particularities of historical documents, such as noise and degradation caused by copying, scanning or aging (e.g. yellow pages, ink stains, mold or moisture, faded out ink, uneven lighting due to folded, corrugated parchment or papyrus). Therefore, applying an enhancement method or a denoising technique is considered as the first major step in a document image analysis workflow. It is well-known that the success or failure of a historical document image analysis and understanding approach tightly depends on the results of applying an enhancement, a denoising or a restoration technique as a pre-processing stage in a historical document image analysis workflow. Several scientific works in historical document image analysis have described several relevant approaches for restoration or enhancement purposes [8].

In the literature, the issue of enhancing and restoring historical document images has been tackled by using different filtering techniques. Filtering techniques have been intensively used in different fields of pattern recognition and machine vision. It has been observed that filtering techniques are flexible in a wide range of image types for reducing or eliminating noise on the one hand, and for improving edge contrast on the other hand [3]. Nevertheless, it is worth noting that the significant increase of image complexity of historical documents is considered one serious limitation for the choice of the appropriate filtering technique [13], [9]. Moreover, applying a filtering technique on historical documents digitized in full color requires an additional processing to handle with the color space. Applying a filter on a color document image should be addressed differently from a gray-scale one [7], [2]. Indeed, applying a filter on color document images involves the use of a specific color model [11]. Each pixel is represented by a multi-dimensional vector that is determined according to the specified color space model such as RGB (red, green and blue), HSB (hue, saturation and brightness) and CMYK (cyan, magenta, yellow and key), *etc.*

Several research works have suggested that the design of a filtering technique for color images should take into account the multi-dimensionality paradigm when examining the theoretical basis for such an approach [6], [12]. For instance, Angulo and Serra [1] presented a theory of multi-dimensional mathematical morphology for color images. To model color image, Ell and Sangwine [5] introduced the theory of vector filters based on linear quaternion functions.

---

[1] http://www.archives.nat.tn/
E-mail addresses: walid.hedda@enit.rnu.tn (W. Elhedda),
maroua.mehri@eniso.u-sousse.tn (M. Mehri),
maroua.mehri@eniso.u-sousse.tn (M.A. Mahjoub)

\* Corresponding author

Applying a filter to color images is tackled in different ways by many researchers. Most researchers have usually applied filters to color images either by analyzing separately each color component of the selected color model or by defining a linear combination of the different color components of the selected color model. In some cases, other researchers have avoided separating the different color components and have proposed to use a hyper-complex or a hyper-spatial model when applying filters to color images. For instance, Sangwine and Ell [10] introduced a filter based on convolution with hyper-complex masks for color images. Nevertheless, when using a hyper-complex or a hyper-spatial model, *a priori* knowledge regarding the content of the analyzed data is absolutely required in order to define the appropriate filters for color images. Therefore, research efforts have recently been turned towards enriching semantically the applied filters in order to adapt them properly to the content of the analyzed data. This may eventually result in a vicious circle between filtering, segmentation and recognition to find the most fine-tuned filter. Among the serious limitations of applying filters on color images that have recently been highlighted by numerous researchers, include first the lack of reliable and accurate segmentation methods of color document images into color layers. For instance, some filtering techniques have recently been proposed based on analyzing only few color components, which may cause significant performance loss. In addition, it is commonly agreed that the key challenge of color image processing lies in finding the adequate color space model, which is not a trivial task. Nevertheless, Elhedda [4] claimed that the choice of the specified color component of the color space model to analyze is a more determining factor than the choice of the color space model for color image segmentation. It is also important to be noted that the type of the filtering approach used for segmenting color images has a significant impact.

Filtering techniques for color images can be categorized according to the adopted type of analysis strategy. There are four categories of approaches for filtering techniques used for color images: scalar, marginal, vector and hybrid. In our work, we only focus on investigating the marginal and vector approaches. The purpose of our work is to provide valuable guidelines on the strengths and weaknesses of these filtering approaches.

The remainder of this article is organized as follows. Section 2 reviews a number of commonly and widely used conventional filtering techniques applied to color images. In Section 3, we outline the criterion used to evaluate the performance of the different assessed filtering techniques in this work. In Section 4, our experimental protocol is firstly detailed. Then, we discuss quantitatively the obtained performance of the filtering technique experiments. Finally, our conclusions and future work are presented in Section 6.

For simplification purposes, we have adopted the following notations in this article.
- $I$ is the input image,
- $F$ is the filtered image,
- $M$ denotes the number of image intensities,
- $N$ corresponds to the number of image pixels.

## 2. Filtering techniques for color images

In the literature, the most known filtering approaches are: scalar, marginal, vector and hybrid. A scalar approach combines the color components, defined according to the selected color space model, into a single numeric value. A marginal approach operates by applying filters separately on each color component determined according to the selected color space model. A vector approach operates by applying filters on vector representations defining the different color component values of the analyzed pixels. A hybrid approach combines several types of filtering approaches. Based on the assumption that color images can best be handled within investigating all available color information, we focus in our work on evaluating the vector and marginal approaches (*i.e.* a scalar approach eliminates the color information) [7], [2].

Filtering techniques are largely used for both denoising and edge detection purposes. It is important to note that the definition of a filter is closely associated with the hyperspace notion. Indeed, some filters need to be reviewed once the properties of a hyperspace are modified. For instance, applying a Laplacien filter in a two-dimensional space only requires a simple convolution of the analyzed image with the Laplacian kernel, while applying a Laplacien filter in a multi-dimensional space needs to review the filter principles (*i.e.* by using a finite-difference method). However, other filters do not need to be re-examined such as the mean filter. The mean filter gives similar results when analyzing each color component separately and when analyzing the hyperspace. Thus, for both denoising (*see.* Section 2.1) and edge detection (*see.* Section 2.2) purposes, a number of commonly and widely used filters and their application in a hyperspace are summarized in the following.

### 2.1 Denoising filters

Denoising filters are designed to reduce or to eliminate different noise levels and degradation types existing in an image. A denoising filter requires *a priori* knowledge regarding the image to be analyzed and their characteristics (e.g. noise levels and types). There is a variety of filters used for denoising purposes. In our work, we only focus on the conventional denoising filters, including the mean filter (*see.* Section 2.1.1), median filter (*see.* Section 2.1.2) and morphological filter (*see.* Section 2.1.3).

### 2.1.1 Mean filter

To compute the intensity value of a pixel by using a mean filter, we need to compute its average intensity value regarding its neighbors. In the case of a color image, the computation of the average intensity value of a pixel regarding its neighbors is based on determining the centroid.

The centroid is given by:

$$\vec{OG} = \frac{\sum_i m_i \cdot \vec{OM_i}}{\sum_i m_i} \quad (1)$$

Where $G$ denotes the gravity center of the point cloud represented by $M_i$, O is the origin of the vectors, and $m_i$ corresponds to the weight of $M_i$.

Subsequently, for color images the coordinates of the point $G$ are calculated separately for each color component of the selected color space model. Consequently, the marginal and vector approaches will have similar results in the case of using the mean filter. Similarly, with regard to the weighted mean filter and Gaussian filter, there will no differences in terms of results when using either a marginal approach or a vector one.

### 2.1.2 Median filter

The median filter is based on sorting the intensity values of the pixel to be analyzed and its neighbors. Afterwards, the median value is assigned to the pixel to be analyzed. In the hyperspace of a color image (*see.* Equation 2), sorting the intensity values of a set of pixels is equivalent to sorting the norms of the intensity vectors (*see.* Equation 3).

$$I(x-1:x+1, y-1:y+1) = \begin{pmatrix} a & b & c \\ d & e & f \\ g & h & i \end{pmatrix} \quad (2)$$

Where {a, b, … h, i} denotes a set of vectors.

Each vector contains the different color components of the selected color space model.

$$F(x,y) = median(||a||, ||b||, \cdots, ||h||, ||i||) \quad (3)$$

### 2.1.3 Morphological filters

The idea of a basic morphological filter is based on sorting filters. For a morphological erosion operation, we simply need to sort the intensity values of the pixel to be analyzed and its neighbors. Afterwards, the minimum intensity value of the sorted set of the intensity values is assigned to the pixel to be analyzed (*see.* Equation 4). In the case of color images, the vector norms are sorted.

$$F(x,y) = minimum(||a||, ||b||, \dots, ||h||, ||i||) \quad (4)$$

Similarly, for a morphological dilation operation the maximum intensity value of the sorted set of the intensity values is assigned to the pixel to be analyzed (*see.* Equation 5).

$$F(x,y) = maximum(||a||, ||b||, \dots, ||h||, ||i||) \quad (5)$$

The morphological erosion or dilation operation is operated by using a 3 x 3 square structuring element on the hyperspace of a color image (*see.* Equation 2).

## 2.2 Edge detection filters

Edge detection is an important pre-processing step preceding a series of image analysis tasks. It is based on identifying points having more or less abrupt variation in terms of image intensities with the aim of capturing and excluding noise. The main edge detection filters are: Roberts filter, Laplacian filter, Sobel filter, Prewit filter, Kirsh filter, Canny filter, Derich filter and Gabor filter. In the following, we outline briefly the use of the Laplacian and the Sobel filters for color images in Sections 2.2.1 and 2.2.2, respectively.

### 2.2.1 Laplacian filter

The Laplacian filter is based on computing the second derivative of the intensity function. It is defined by means of a finite-difference method computed at the pixel $I(x,y)$.

The kernel of the Laplacian filter is given by:

$$\begin{pmatrix} 0 & 1 & 0 \\ 1 & -4 & 1 \\ 0 & 1 & 0 \end{pmatrix} \quad (6)$$

Applying a Laplacian filter on a gray-scale image requires a simple convolution of the analyzed image with the Laplacian kernel. The Laplacien filter remains a useful technique for color images in the case of using a marginal approach. However, applying a Laplacian filter with a marginal approach (*i.e.* each color component is separately analyzed) is really incorrect due to the fact that an edge can not be dislocated on several planes. Hence, the idea is to use a vector approach. In this case, a pixel is characterized by an intensity vector.

The Laplacian filter applied for color images is computed as follows:

$$F(x,y) = \sqrt{LR(x,y) + LG(x,y) + LB(x,y)} \quad (7)$$

Where

$$\begin{aligned}LR(x,y) = &(I_R(x-1,y) - I_R(x,y))^2 + (I_R(x,y-1) \\ &- I_R(x,y))^2 + (I_R(x,y+1) \\ &- I_R(x,y))^2 + (I_R(x+1,y) \\ &- I_R(x,y))^2\end{aligned}$$

$$\begin{aligned}LG(x,y) = &(I_G(x-1,y) - I_G(x,y))^2 + (I_G(x,y-1) \\ &- I_G(x,y))^2 + (I_G(x,y+1) \\ &- I_G(x,y))^2 + (I_G(x+1,y) \\ &- I_G(x,y))^2\end{aligned}$$

$$LB(x,y) = (I_B(x-1,y) - I_B(x,y))^2 + (I_B(x,y-1) - I_B(x,y))^2 + (I_B(x,y+1) - I_B(x,y))^2 + (I_B(x+1,y) - I_B(x,y))^2$$

$I_R$, $I_G$ and $I_B$ are the three color components of the analyzed image. This formalism remains consistent for any color space model such as RGB, HSB, *etc*.

### 2.2.2 Sobel filter

A Sobel kernel is applied by means of either a marginal approach or a vector one for edge detection purposes. The only difference between the results of the two approaches when using a Sobel filter lies only in the intensity level of the filtered pixels. The filtered image pixels have higher intensity levels (*i.e.* clearer filtered images) in the case of using a vector approach than in the case of using a marginal one. This is due to the fact that the filtered image is computed by using a vector norm (*i.e.* sum of norm differences). On the other side, in the case of a marginal approach, the sum of differences between pixels is firstly computed and the norm is subsequently calculated.

The kernels of a Sobel filter are defined as follows:
$$S_h(x,y) = \begin{pmatrix} -1 & 0 & 1 \\ -2 & 0 & 2 \\ -1 & 0 & 1 \end{pmatrix} \quad (8)$$

$$S_v(x,y) = \begin{pmatrix} -1 & -2 & -1 \\ 0 & 0 & 0 \\ 1 & 2 & 1 \end{pmatrix} \quad (9)$$

Where $S_h$ and $S_v$ are the horizontal and vertical Sobel kernels, respectively.

The Sobel filter applied for color images by means of a marginal approach is computed as follows:
$$F = \sqrt{F_h^2 + F_v^2} \quad (10)$$

Where $F_h$ and $F_v$ are the resulting images of convoluting the analyzed image with the horizontal and vertical Sobel kernels, respectively.

The Sobel filter applied for color images by means of a vector approach is computed as follows:
$$F(x,y) = \sqrt{SR(x,y) + SG(x,y) + SB(x,y)} \quad (11)$$

Where
$$SR(x,y) = (c_R - a_R)^2 + 2(f_R - d_R)^2 + (i_R - g_R)^2 + (g_R - a_R)^2 + 2(h_R - b_R)^2 + (i_R - c_R)^2$$

$$SG(x,y) = (c_G - a_G)^2 + 2(f_G - d_G)^2 + (i_G - g_G)^2 + (g_G - a_G)^2 + 2(h_G - b_G)^2 + (i_G - c_G)^2$$

$$SB(x,y) = (c_B - a_B)^2 + 2(f_B - d_B)^2 + (i_B - g_B)^2 + (g_B - a_B)^2 + 2(h_B - b_B)^2 + (i_B - c_B)^2$$

$$\begin{pmatrix} a_\alpha & b_\alpha & c_\alpha \\ d_\alpha & e_\alpha & f_\alpha \\ g_\alpha & h_\alpha & i_\alpha \end{pmatrix} = \begin{pmatrix} I_\alpha(x-1,y-1) & I_\alpha(x-1,y) & I_\alpha(x-1,y+1) \\ I_\alpha(x,y-1) & I_\alpha(x,y) & I_\alpha(x,y+1) \\ I_\alpha(x+1,y-1) & I_\alpha(x+1,y) & I_\alpha(x+1,y+1) \end{pmatrix}$$

We note that $\alpha$ denotes either R, G and B for RGB space model or H, S and B for HSB space model. This formalism remains consistent for other color space models.

## 3. Performance evaluation criteria

In order to assess the different filtering techniques, previously presented in Section 2, several assessment criteria for filter performance evaluation are detailed in this section. Two types of assessment criteria can be used to evaluate filter performance. The first type of assessment criteria is psycho-visual. The second type of assessment criteria is quantitative or statistical. Comparing visually the effectiveness of a filtering technique is inherently a subjective evaluation and is not sufficient. Thus, we focus on the second type of assessment criteria in this work. Indeed, the performance of the different filtering techniques has been analyzed by computing three metrics including: the peak signal-to-noise ratio (*PSNR*), the ratio of the regions statistics (*SR*) and the statistical variance of a homogeneous region in an image ($R_{SC}$) [12]. These metrics depends on the type of the used filtering technique.

- The *PSNR* metric (*see.* Section 3.1) and the *SR* metric (*see.* Section 3.2) are computed in the case of using a denoising filter,
- The $R_{SC}$ metric is computed in the case of using an edge detection filter (*see.* Section 3.3).

### 3.1 *PSNR* metric

The peak signal-to-noise ratio (*PSNR*) is computed as follows:
$$PSNR = 10 \log_{10} \frac{Max^2}{MSE} \quad (12)$$

Where
- *MSE* denotes the mean square error, which is given by:
$$MSE = \sum_x \sum_y \frac{(I(x,y) - F(x,y))^2}{N} \quad (13)$$

- *Max* denotes the maximum value of image intensity.

This is exemplified by the following calculation method applied to RGB color image by using a vector form. A vector form is used to determine $(I(x,y) - F(x,y))^2$ which is giving by:

$$(I(x,y) - F(x,y))^2 = (I_R(x,y) - F_R(x,y))^2 + (I_G(x,y) - F_G(x,y))^2 + (I_B(x,y) - F_B(x,y))^2 \quad (14)$$

The higher the values of the computed *PSNR* criterion, the better the filter results.

## 3.2 *SR* metric

The ratio of region statistics (*SR*) is computed by:
$$SR = \frac{\sum_i \sigma_i}{m \cdot N} \quad (15)$$

Where
- $\sigma_i$ denotes the standard deviation of the i[th] region,
- *m* is the statistical mean of the whole image intensities.

The regions are identified by convoluting an image difference (between the original image and the filtered image) at a window having a *T* x *T* size (*T* is equal to 5).

The lower the values of the computed *SR* criterion, the better the filter results.

## 3.3 $R_{SC}$ metric

The $R_{SC}$ metric is computed based on a statistical analysis of image regions. It uses the Lee filter to crop an image into regions and subsequently statistical moments for each image regions are calculated. For each region, two simple statistics, the mean and standard deviation, deduced from the variations of each edge of the analyzed pixel are computed. The $R_{SC}$ criterion corresponds to the ratio between the sum of the standard deviations applied to the mean difference norm of image regions of either side of the edge. The mean and standard deviation are computed twice, either side of the edge.

Usually, four scan directions have been used: $\theta_r = \{0°, 45°, 90°, 135°\}$ (*i.e.* horizontal, vertical, diagonal and anti-diagonal directions).

Figure 1 illustrates the four scan directions used for computing the statistics of the $R_{SC}$ metric. Therefore, statistics are computed for each scan direction.

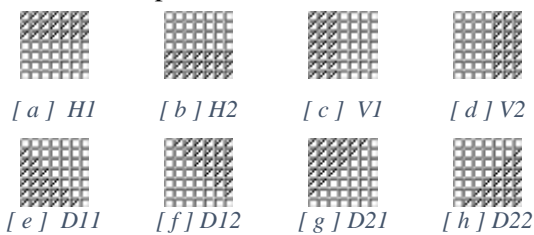

[ a ] H1    [ b ] H2    [ c ] V1    [ d ] V2

[ e ] D11    [ f ] D12    [ g ] D21    [ h ] D22

*Figure 1: Illustration of the four scan directions used for computing the statistics of the $R_{SC}$ criterion: horizontal (H1 and H2), vertical (V1 and V2), first diagonal from the right to the left (D11 and D12), second diagonal from the left to the right (D21 and D22).*

The $R_{SC}$ criterion is given by:

$$R_{SC} = \sum_x \sum_y \frac{\sqrt{\sum_z r_z^2}}{N} \quad (16)$$

Where
- z is the specified direction (*i.e.* horizontal, vertical, diagonal and anti-diagonal directions).
- $r_z$ denotes the ratio in the z direction. $r_z$ is defined as:
$$r_z = \frac{|m_{z1} - m_{z2}|}{\sqrt{v_{z1} + v_{z2}}} \quad (17)$$
- $m_{z1}$ and $v_{z1}$ correspond to the mean and variance, respectively, computed according to the predefined edge side (*see.* Figure 1) in the specified direction.

The higher the values of the computed $R_{SC}$ criterion, the better the filter results.

## 4. Evaluation results

To illustrate the effectiveness of the different assessed filtering techniques, a thorough experimental study has been conducted in this article using a corpus of color archival document images collected from the Tunisian national archives. In the following, we firstly detail our experimental protocol and the experiments carried out to evaluate different filtering techniques applied to color archival document images (*see.* Section 4.1). Then, we discuss quantitatively the obtained filter performance (*see.* Section 4.2).

### 4.1 Experimental protocol

The denoising filters evaluated in this work are: the median filter with a kernel having a 3 x 3 size, and the morphological filter having the following structure "opening (closing (image)) ". Each filter is applied using the following approaches: a marginal approach (*M.*), a vector approach (*V.*), and two dual approaches. A dual approach is considered as a kind of a hybrid one. In our work, we have evaluated two dual approaches. The first dual approach (*MV.*) is a succession of marginal and vector approaches, while the second one (*VM.*) is a succession of vector and marginal approaches.

The experimental corpus used in our work has been collected from the Tunisian national archives. Figure 2 illustrates an example of an archival document image collected from the Tunisian national archives.

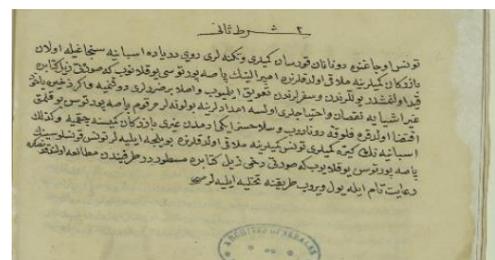

*Figure 2. An example of an archival document image collected from the Tunisian national archives.*

First, the different filtering techniques assessed in our work have been applied without introducing any noise on images of our experimental corpus. Then, several degradations are added to images with the goal of evaluating the robustness of the filtering techniques on color images. Therefore, images under numerous degradation and noise models have been generated in order to evaluate the performance of the different filtering techniques and approaches.

Several degradation and noise models have been introduced by adding several kinds of noise on images of our experimental corpus. The different noise models assessed in our work are presented in Table 1.

*Table 1. List of the different evaluated noise models.*

| *Id.* | Description |
|---|---|
| *Noise 1* | Weak additive Gaussian noise |
| *Noise 2* | Strong additive Gaussian noise |
| *Noise 3* | Weak speckle noise |
| *Noise 4* | Strong speckle noise |
| *Noise 5* | Weak "salt & pepper" noise |
| *Noise 6* | Strong "salt & pepper" noise |

The filters assessed for edge detection purposes are: the Laplacian filter, Sobel filter and morphological filter. The morphological filter is obtained by computing the difference between the image filtered by means of a morphological dilatation and the image filtered by means of morphological erosion.

### 4.2 Results

The performance evaluation and comparison of the different filtering techniques used for denoising and edge detection purposes, are presented in Sections 4.2.1 and 4.2.2, respectively.

#### 4.2.1 Denoising filters

Figures 5 and 6 illustrate the obtained performances of the median and morphological filters, respectively, in the RGB color space by computing the *PSNR* metric.

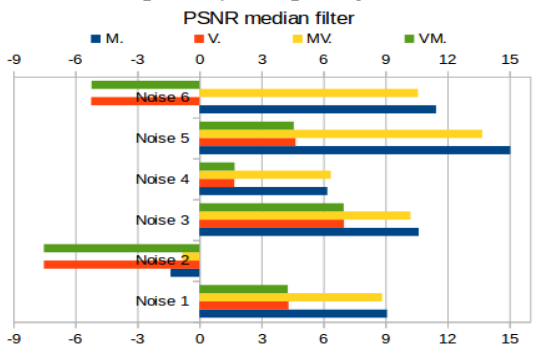

*Figure 3. Performance evaluation of the median filter in the RGB color space by computing the PSNR metric.*

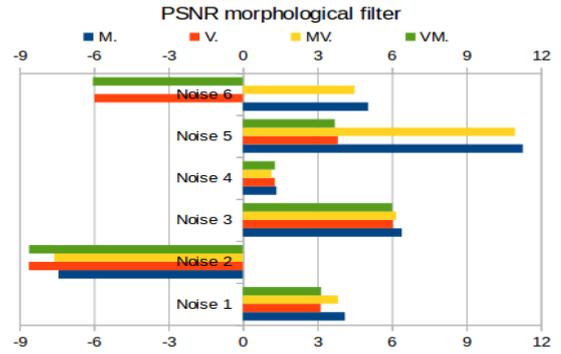

*Figure 4. Performance evaluation of the morphological filter in the RGB color space by computing the PSNR metric.*

By computing the PSNR metric, we note that the best performance is obtained by using a marginal approach for almost all types of noise (*see*. Figures 3 and 4).

Figures 5 and 6 illustrate the obtained performances of the median and morphological filters, respectively, in the RGB color space by computing the *SR* metric.

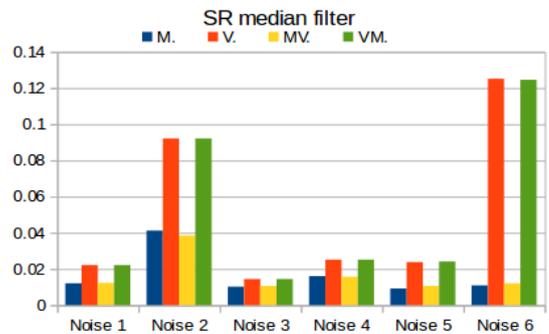

*Figure 5. Performance evaluation of the median filter in the RGB color space by computing the SR metric.*

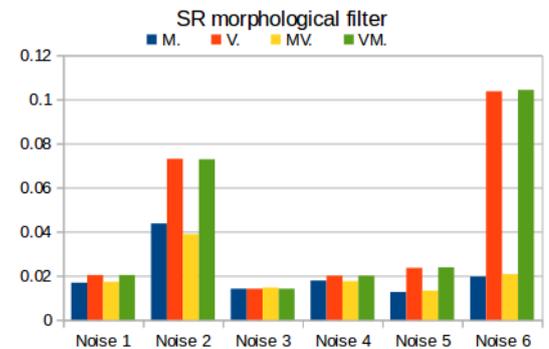

*Figure 6. Performance evaluation of the morphological filter in the RGB color space by computing the SR metric.*

Similarly, by computing the *SR* metric we also observe that a marginal approach outperforms the other evaluated approaches for almost all types of noise either when using the median filter or the morphological filter (*see*. Figures 5 and 6). This strengthens our previous observations obtained when analyzing the obtained *PSNR* results. We also observe that the two assessed dual approaches (*MV.* and *VM.*) do not perform better than the other evaluated approaches in most of carried out experiments for image denoising purposes. This can be justified by the produced bias by using a progressive merge process of two different filtering approaches.

Figures 7 and 8 illustrate the obtained *PSNR* and *SR* performances, respectively by using the median and morphological filters.

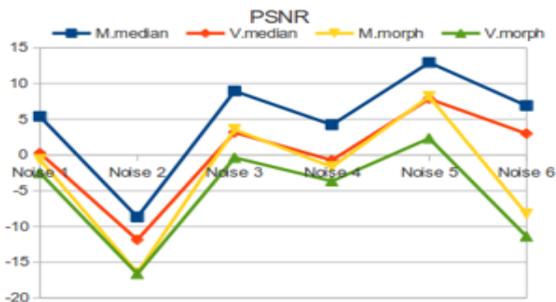

*Figure 7. Performance evaluation of the median and morphological filters in the HSB color space by computing the PSNR metric. The median and morphological filters are annotated with "median" and "morph", respectively.*

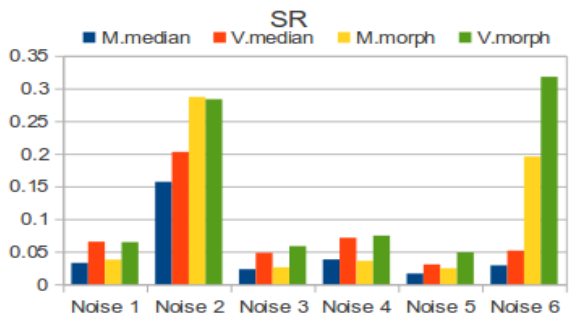

*Figure 8. Performance evaluation of the median filter and the morphological filter in the HSB color space by computing the SR metric. The median and morphological filters are annotated with "median" and "morph", respectively.*

We note that the experiments carried out on other color spaces such as HSB (*see*. Figures 7 and 8) confirms our previous observations concerning the RGB color space (*see*. Figures 3, 4, 5 and 6). Indeed, similar performance is obtained according to the most computed evaluation metrics when comparing results obtained from two different color spaces.

### 4.2.2   Edge detection filters

Figure 9 depicts the obtained performances of the morphological, Sobel and Laplacian filters by computing the $R_{SC}$ metric.

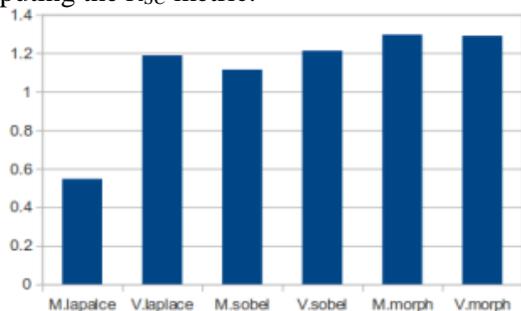

*Figure 9. Performance evaluation of the morphological, Sobel and Laplacian filters by computing the $R_{SC}$ metric. The morphological, Sobel and Laplacian filters are annotated with "laplace", "sobel" and "morph", respectively.*

We note that using a vector approach gives better results for the Laplacian and Sobel filters. This can be explained by the limitations of a marginal approach to split spatially edge in many hyper-space plans. Nevertheless, the results of the morphological filter are similar when using vector and marginal approaches (*see*. Figure 9).

## 5. Conclusions and further work

The filtering of color images have many particularities compared to the filtering of gray-scale images. Indeed, the filtering techniques in gray levels are only based on computing pixels intensities. On the other side, analyzing color images is a tricky task. Hence, applying a filtering technique on a color image requires more extensive image processing tasks to handle with the color space. The major difficulty encountered at filtering color images lies in the number of intensities contained in a single pixel.

Considering the type of the adopted approach in the filtering technique, each filter is analyzed according to their characteristics and performance. For instance, the conventional linear filters, such as the mean and Gaussian filters, make a vector approach similar to a marginal one. In the case of using high-order filters, such as the median and morphological filters, a marginal approach is better than a vector one. This can be justified by the need to take into account the characteristics of the existing noise in the analyzed data. Indeed, if we have *a priori* knowledge about the noise and the image content, a vector approach will outperform a marginal one. Therefore, the choice of a filtering approach tightly depends on the availability of *a priori* knowledge on the image characteristics. Hence, a marginal approach is the suitable choice when applying a "blind" filtering on a color image (*i.e.* without *a priori* knowledge of the image characteristics).

Based on our experimental results and observations, we conclude that adopting a marginal approach in using a filtering technique that do not require *a priori* knowledge of the characteristics of data to be analyzed is the best alternative for enhancing or denoising historical document images. Nevertheless, adopting a vector approach in using a filtering technique for edge detection purposes remains the suitable choice.

Our further works will be in line with those described in this article. The first aspect of future work will be to extend our investigation to other datasets. Furthermore, we intend to conduct a comparative study of the quaternion and vector filters. Finally, we will investigate filters in the fifth space dimension or more.

## References

[1]   Angulo, J., and Serra, J., "Color segmentation by ordered mergings," in International Conference on Image Processing, pp. 14-17, 2003.

[2]   Drira, F., Lebourgeois, F., and Emptoz, H., "Restoring Ink Bleed-Through Degraded Document Images Using a Recursive Unsupervised Classification Technique," in International

Workshop on Document Analysis Systems, pp. 38-49, 2006.

[3] Eglin, V., Bres, S., and Rivero, C., "Hermite and Gabor transforms for noise reduction and handwriting classification in ancient manuscripts," *International Journal of Document Analysis and Recognition*, vol. 9, no. 2-4, pp. 101-122, 2007.

[4] Elhedda, W., "Color space choice for color images segmentation," in Colloque de la Recherche Appliquée et de Transfert de Technologie, 2009.

[5] Ell, T. A., and Sangwine, S. J., "Theory of vector filters based on linear quaternion functions," in European Signal Processing Conference, pp. 1-5, 2008.

[6] Gevers, T., Gijsenij, A., Vandeweijer, J., and Geusebroek, J. M., *Color in Computer Vision: Fundamentals and Applications*. John Wiley & Sons, 2012.

[7] Ghomrassi, A., Charrada, M. A., and Amara, N. E. B., "Restoration of ancient colored documents, foreground/background separation," in International Multi-Conference on Systems, Signals and Devices, pp. 1-6, 2015.

[8] Mehri, M., Sliti, N., Héroux, P., Gomez-Krämer, P., Amara, N. E. B., and Mullot, R., "Use of SLIC superpixels for ancient document image enhancement and segmentation," in Document Recognition and Retrieval, vol. 9402, 2015.

[9] Rana, B., Kapoor, N., and Kundra, H., "A review on the image enhancement of old historical documents and images," *International Journal of Advances in Science and Technology*, vol. 4, no. 1, pp. 180-185, 2016.

[10] Sangwine, S. J., and Ell, T. A., "Colour image filters based on hypercomplex convolution," *IEEE Proceedings - Vision, Image and Signal Processing*, vol. 147, no. 2, pp. 89-93, 2000.

[11] Sgarbi, E. M., Mura, W. A. D., Moya, N. and Facon, J., and Ayala, H. A. L., "Restoration of old document images using different color spaces restoration of old document images," in International Conference on Computer Vision Theory and Applications, pp. 82-88, 2014.

[12] Vandenbroucke, N., Macaire, L., and Postaire, J. G., "Color Pixels Classification in an Hybrid Color Space," in International Conference on Image Processing, pp. 176-180, 1998.

[13] Yahya, S. R., Abdullah, S. N. H. S., Omar, K., Zakaria, M. S., Liong, C. Y., "Review on image enhancement methods of old manuscript with damaged background," in International Conference on Electrical Engineering and Informatics, pp. 62-67, 2009.

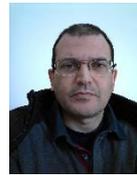

**Walid Elhedda**

Walid Elhedda has a Bachelor.Sc. in Electrical Engineering in 1995, Practical Master in Electrical Engineering in 1998 and Master.Sc. in Automatic and Signal Processing in 2004 from the National Engineering School of Tunis (ENIT). Currently, he is an assistant professor at Sousse University and a member of the Laboratory of Advanced Technology and Intelligent Systems (LATIS).

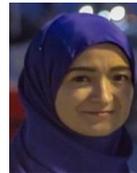

**Maroua Mehri**

Maroua Mehri has a Ph.D. in computer science from the University of La Rochelle in 2015. She has an engineering degree in computer engineering from the National Engineering School of Sousse (ENISo) in 2009 and two MSc. degrees in image processing from the University of Rennes 1 and the ENISo in 2011. She is currently an assistant professor in computer science at the Computer Engineering Department in ENISo. Her research activities are currently carried out at the Laboratory of Advanced Technology and Intelligent Systems (LATIS). Her main research interests are pattern recognition and machine learning in the field of historical document image analysis.

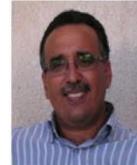

**Mohamed Ali Mahjoub**

Mohamed Ali Mahjoub is an associate professor in Signal and Image processing at the National Engineering School of Sousse (ENISo) and member of the Laboratory of Advanced Technology and Intelligent Systems (LATIS). His research interests include dynamic bayesian network, computer vision, pattern recognition, HMM and data retrieval. He is a member of IEEE and his main results have been published in international journals and conferences.